\documentclass[10pt, letter, journal]{IEEEtran}         

\usepackage{amsmath,amssymb,amsfonts}
\usepackage{graphicx}
\usepackage{pifont}

\usepackage{amsthm}
\usepackage{xcolor}
\usepackage[dvipsnames]{xcolor}
\usepackage[figuresright]{rotating}

\usepackage{graphicx}
\graphicspath{{/}{fig/}}
\usepackage{graphicx}
\usepackage{array}
\usepackage{textcomp}
\usepackage{booktabs}

\usepackage{mathtools}
\usepackage{breqn}
\usepackage{float}

\usepackage{pgfplots}
\pgfplotsset{compat=1.7}
\usepgfplotslibrary{groupplots}

\usepackage{caption}
\usepackage{subcaption}

\usepackage{multirow,tabularx}
\usepackage{hyperref}
\usepackage{flushend}
\usepackage{algorithmic}
\usepackage[vlined, ruled, shortend]{algorithm2e}

\usepackage{cleveref}

\newcommand{\etal}{\textit{et al.}}

\newlength\figureheight
\newlength\figurewidth
\SetAlCapNameFnt{\footnotesize}
\SetAlCapFnt{\footnotesize}

\crefname{figure}{Fig.}{Figs.}
\Crefname{figure}{Fig.}{Figs.}
\Crefname{section}{Section}{Section}
\crefname{section}{Section}{Section}
\Crefname{table}{Table}{Table}
\crefname{table}{Table}{Table}

\title{
    EgoMoD: Predicting Global Maps of Dynamics from Local Egocentric Observations
}

\author{
    Iacopo Catalano, David Morilla-Cabello, Jorge Peña-Queralta, Eduardo Montijano
    \thanks{This work was partially supported by the Kaute Foundation through the Tutkijat Maailmalle program and by grants AIA2025-163563-C31, PID2024-159284NB-I00, funded by MCIN/AEI/10.13039/501100011033 and ERDF, the Office of Naval Research Global grant N62909-24-1-2081 and DGA project T45\_23R.}
    \thanks{I. Catalano is with the University of Turku, Finland.}
    \thanks{J. Pe\~na-Queralta is with the Centre for Artificial Intelligence, Zürich University of Applied Sciences, Winterthur, Switzerland.}
    \thanks{E. Montijano and D. Morilla-Cabello are with the Instituto de Investigaci\'on en Ingenier\'ia de Arag\'on, Universidad de Zaragoza, Spain.}
    \thanks{Corresponding: \texttt{\small imcata@utu.fi}}
    \thanks{The code will be released upon acceptance}
}

\begin{document}

\maketitle
\thispagestyle{empty}
\pagestyle{empty}

\begin{abstract}%
    \label{sec:abstract}%
    Efficient navigation in dynamic environments requires anticipating how motion patterns evolve beyond the robot's immediate perceptual range, enabling preemptive rather than purely reactive planning in crowded scenes.
    Maps of Dynamics (MoDs) offer a structured representation of motion tendencies in space useful for long-term global planning, but constructing them traditionally requires global environment observations over extended periods of time.
    We introduce EgoMoD, the first approach that learns to predict future MoDs directly from short egocentric video clips collected during robot operation.
    Our method learns to infer environment-wide motion tendencies from local dynamic cues using a video- and pose-conditioned architecture trained with MoDs computed from external observations as privileged supervision, allowing local observations to serve as predictive signals of global motion structure.
    Thanks to this, we offer the capacity to forecast future motion dynamics over the whole environment rather than merely extend past patterns in the robot's field of view.
    As a site-specific dynamic prior, EgoMoD replaces the external global sensing infrastructure required by prior MoD methods at inference time with standard onboard sensors.
    Experiments in large simulated environments show that EgoMoD predicts future MoDs under limited observability, while evaluation with real images showcases its zero-shot transferability to real systems.
\end{abstract}
\IEEEpeerreviewmaketitle

\section{Introduction}\label{sec:introduction}

Mobile robots operating in populated environments must plan in the presence of dynamic human activity~\cite{singamaneni2024socialsurvey}. Navigation is not only defined by static obstacles but also affected by human movement. These dynamics are often structured in recurrent patterns such as periodic flows in hallways or converging motions around shared facilities. Standard navigation systems, however, typically treat people as independent moving obstacles and respond reactively to immediate observations. This reactive paradigm aims at solving local avoidance but remains myopic to broader scene dynamics that could be anticipated from prior observations of motion patterns. Spatial understanding of collective motion could enable robots to anticipate congestion, identify favorable routes, and plan more efficiently. The central challenge is how to obtain such global motion patterns from the limited egocentric observations a robot naturally collects during operation.

Maps of Dynamics (MoDs)~\cite{kucner2023survey} provide structured models of where and how motion typically occurs, enabling robots to anticipate dynamics beyond their immediate perception. For example, CLiFF-Map \cite{kucner2017enabling} models velocity statistics across spatial regions using a probabilistic framework, while FreMEn-based methods~\cite{krajnik2014spectral, molina2021robotic} capture temporal variations by fitting periodic functions to long-term observations.

Despite their effectiveness, the applicability of these methods for robotic navigation scenarios presents two main limitations.
First, existing MoD algorithms build predictions by fitting statistical or periodic models to accumulated historical data about the dynamics. While effective at capturing recurrent patterns, using only past data prevents the model from incorporating real-time context to anticipate future behaviors.
Second, these models assume constant access to global observations of the environment. Whereas in practice, the observations available to a mobile robot are inherently limited by its line of sight, raising the question of whether it is possible to infer global motion patterns from local viewpoints.

\begin{figure}[t]
         \centering
         \includegraphics[width=0.9\linewidth]{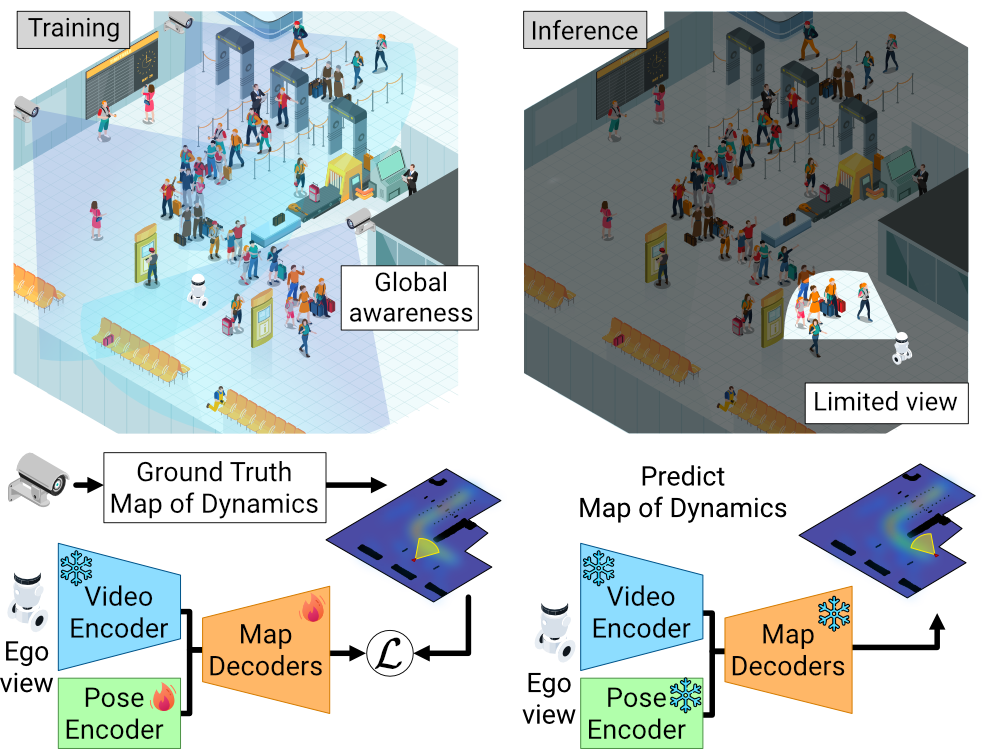}
         \label{fig:mod_office}
    \caption{EgoMoD learns to predict global maps of dynamics (bottom) from local video observations (top) from the egocentric vision of the robot, leveraging priors from a privileged expert.}
    \label{fig:egomod}
    \vspace{-10pt}
\end{figure}

Our main contribution, EgoMoD, is a novel algorithm that addresses these challenges by learning to predict future and global MoDs from current and egocentric videos. 
We propose a new architecture that extract visual temporal representations from short clips using video foundation models~\cite{assran2025v}, combines them with pose information and decodes them into Bird’s-Eye-View (BEV) grids that anticipate collective motion patterns in the whole environment.
Importantly, the model is supervised using future MoDs of the scene constructed from external global observations (see~\cref{fig:egomod}, left), enabling the system to transform local observations into global predictions, producing forecasts not only in directly observed regions but also in unobserved areas (see~\cref{fig:egomod}, right).
At inference time, we only require standard onboard sensors such as an RGB camera and localization, making it deployable in realistic robotic settings.
EgoMoD is designed as a site-specific dynamic prior: like prior MoD methods, it is trained for a known deployment environment. The key advance is replacing the external global sensing infrastructure those methods require at inference time with standard onboard sensors, enabling practical deployment without persistent external monitoring.
Results in simulated environments confirm accurate future MoD prediction, while real-image tests highlight zero-shot transfer to real systems.

\section{Related Work} \label{sec:related_work}

Our work builds on research in Maps of Dynamics (MoDs), vision-based perception, and trajectory prediction. In this section, we organize existing approaches according to their input requirements, spatial representations, and learning paradigms, highlighting the design space in which our method operates.

\subsection{Maps Of Dynamics}

MoDs represent spatial motion patterns that enable robots to anticipate where and when movement typically occurs \cite{kucner2023survey}. Existing approaches fundamentally differ in what types of input data they require and how they model spatial understanding. CLiFF-Map \cite{kucner2017enabling} relies on privileged velocity measurements collected across spatial regions, building probabilistic models of speed and orientation patterns. 

Temporal dynamics modeling represents another key dimension where approaches diverge. FreMEn \cite{krajnik2017fremen} and its variants \cite{krajnik2014spectral, molina2021robotic, vintr2019time} conceptualize environments as temporally varying rather than static, learning periodic patterns from long-term observations. Recent efforts \cite{shi2023learning, shi2025learning} further extend this concept by learning how spatial motion patterns themselves evolve over time, representing gradual rather than purely periodic changes in movement behaviors.
Continuous representations have also been proposed to overcome the limitations of grid-based MoDs using implicit neural representations~\cite{zhu2025neural} or Gaussian Process models~\cite{stuede2022non}.
In another direction, \cite{verdoja2024bayesian} incorporates static environmental geometry to inform dynamic predictions, relying on external sensing for accurate motion modeling. Yan~\etal~\cite{yan2025stef} combine periodic spectral models with LSTM networks for long-term MoD prediction, while event-triggered frameworks~\cite{shi2025event} employ neural stochastic differential equations to capture changes in motion patterns.

By leveraging these methods for supervision, our approach enables inference of spatial global motion patterns directly from short egocentric video sequences collected during regular robot operation, alleviating the need for external monitoring infrastructure at deployment time.

\subsection{Trajectory Forecasting}

Trajectory forecasting focuses on predicting future motion based on historical position data. Early methods, such as  Conditional Variational Autoencoder (CVAEs) \cite{salzmann2020trajectron++} and  Long
Short-Term Memory (LSTM) \cite{alahi2016social}, have evolved into sophisticated Transformer-based and diffusion models \cite{yuan2021agentformer, gu2022stochastic} that account for social interactions between agents. Recent work has shifted toward joint forecasting \cite{samavi2025sicnav}, predicting both human and robot trajectories to optimize robot motion based on the anticipated movement of others.

Complementary approaches address crowd dynamics through different modeling paradigms. Flow-based methods model crowd dynamics as continuous pseudo-fluid fields with viscosity constraints \cite{dugas2022flowbot}, but require external LiDAR sensing for real-time navigation planning. Normalizing flows with RNN autoencoder abstractions capture full trajectory distributions by learning over abstracted trajectory features, but focus on short-term individual prediction \cite{meszaros2024trajflow}. 
Macroscopic approaches for crowd modeling~\cite{kiss2021probabilistic}, focus on aggregating crowd behaviors rather than individual trajectories. 
Recent work has also shown that MoDs can improve long-term trajectory forecasting when used as priors to bias velocity sampling~\cite{zhu2025long}, and that integrating spatial motion patterns from MoDs into diffusion-based generative models enhances cross-domain adaptability~\cite{shi2025leveraging}.

Unlike trajectory forecasting methods, which predict the future paths of specific agents, our approach forecasts aggregate motion tendencies without requiring explicit agent tracking at inference time. It therefore models collective spatial dynamics rather than individual trajectories, while remaining compatible with egocentric sensing.

\subsection{Egocentric Visual Perception and Spatial Reasoning}

While trajectory forecasting addresses motion prediction from position data, learning spatial understanding from egocentric visual observations presents a distinct challenge. Asghar \etal~\cite{asghar2022allo} explore how allocentric representations improve motion prediction in dynamic scenarios, while Bigazzi \etal~\cite{bigazzi2024mapping} use CLIP-based region mapping to build global semantic maps from RGB-D observations.

Recent end-to-end neural approaches further explore spatial learning from egocentric observations, including variational frameworks for occupancy prediction \cite{lu2019monocular}, pyramid networks for BEV mapping \cite{roddick2020predicting}, spatial anticipation beyond visible regions \cite{ramakrishnan2020occupancy}, and camera-based future instance prediction \cite{hu2021fiery}. A similar approach to ours uses Convolutional Neural Networks (CNNs) trained on simulated pedestrian data to predict human occupancy distributions, but from static building maps alone, learning transferable walking patterns without requiring any sensor observations \cite{doellinger2018predicting}.

These approaches primarily address geometric occupancy or scene reconstruction, while our method focuses on predicting dynamic motion tendencies.
This distinction highlights the shift from spatial layouts to forecasting collective movement patterns, derived directly from egocentric observations.
\section{Methodology}

\subsection{Problem Setup}

We consider a mobile robot operating in a known populated environment with humans. The robot pose at time $t$ is denoted by $\mathbf{r}^{}_t$ and assumed to be known. The robot is equipped with a camera to obtain a sequence of observations $\mathcal I^{}_t = \left[\mathbf{I}^{}_t, \dots, \mathbf{I}^{}_{t+n}\right]$ where $n$ is the video clip duration. The goal is to predict, using only the local data from the robot, global dynamics of the environment represented as a MoD, $\mathcal{M}$, at a future time $t+T$. Importantly, since our goal is to forecast, we require $n \ll T$.

Our approach operates in two distinct phases. During \textit{training}, we assume the availability of privileged global information from, e.g., external cameras or different robots that observe the scene from complementary viewpoints. These global observations are used to construct ground-truth MoDs, providing supervision that covers the full environment. At \textit{inference}, only the robot's onboard camera and its global position are required (see~\cref{fig:egomod}).

\subsection{Map of Dynamics}
\label{sec:histogram_mod}

To model the environment dynamics, we consider a MoD formulation inspired by~\cite{molina2021robotic}.
MoDs are empirical representations of motion patterns, computed by accumulating detection statistics over time.
Our approach focuses on three relevant dynamic quantities:
\begin{itemize}
    \item \textbf{Flow Magnitude ($\mathcal{F}$):} the overall frequency of motion observed in a given location.  
    \item \textbf{Dominant Direction ($\mathcal{D}$):} the most frequent direction of motion.  
    \item \textbf{Directional Entropy ($\mathcal{E}$):} the variability of motion directions, indicating how diverse movement is in that cell. 
\end{itemize}
Together, dominant direction indicates where the flow is heading, and directional entropy quantifies how reliable that dominant direction is: low entropy signals a well-defined flow direction, while high entropy signals multi-directional or inconsistent flow where a robot should exercise caution regardless of the dominant direction.
Note that these quantities characterize motion patterns at fixed locations in the scene, independently of individual agents.

\begin{figure}[t]
    \centering
    \includegraphics[width=\linewidth]{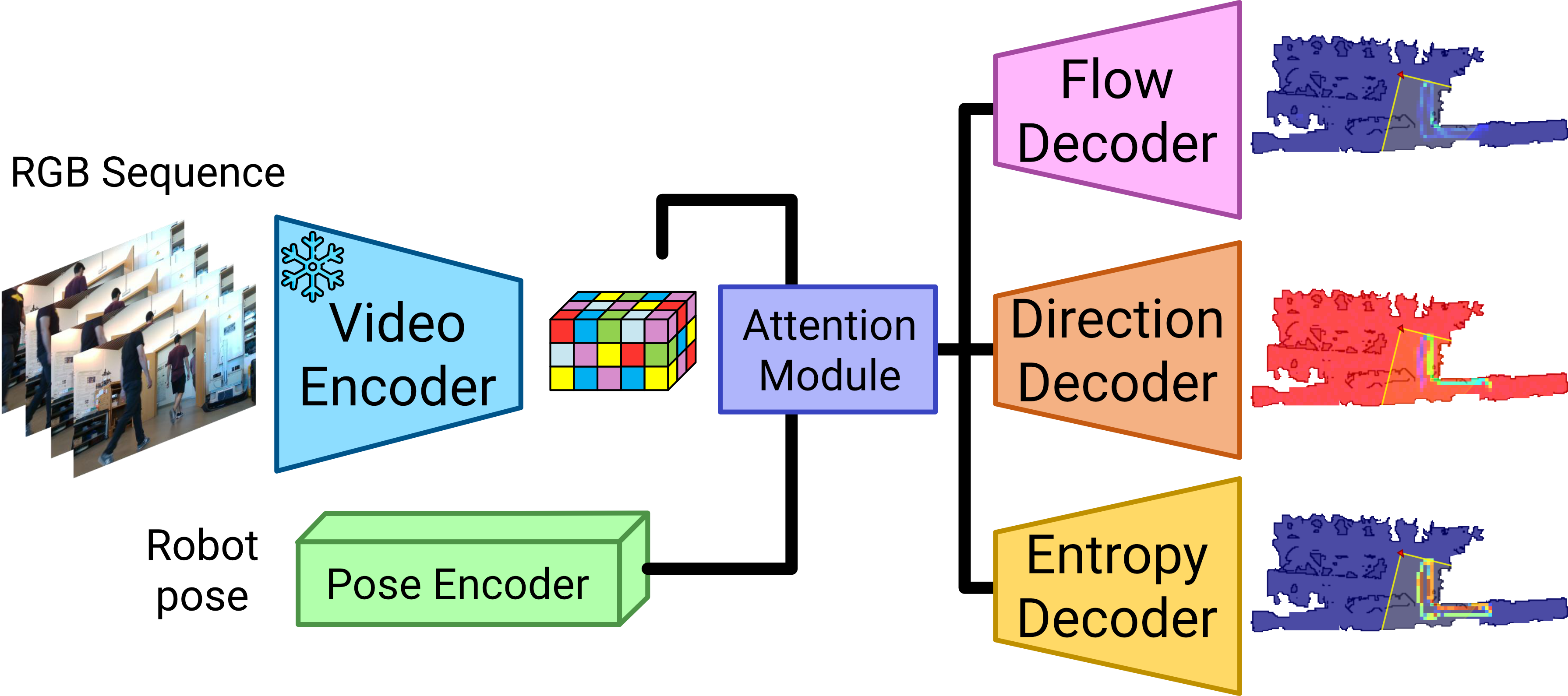}
    \caption{EgoMoD architecture overview. A short egocentric video sequence is processed by a frozen foundation video encoder to extract spatio-temporal features. These features are combined with a learned pose embedding via a Transformer encoder, reshaped to a spatial grid and decoded through convolutional and upsampling layers to produce allocentric Maps of Dynamics.}
    \label{fig:pipeline_diagram}
\end{figure}

\subsection{EgoMoD}

We propose EgoMoD, a supervised learning framework for predicting MoD maps from egocentric visual input. EgoMoD comprises three key components: (i) a neural architecture that uses video and pose information to predict allocentric motion descriptors, (ii) a supervision strategy that leverages privileged global information during training to construct dense MoD maps, and (iii) a set of loss functions that enables training for each of the dynamic quantities.
In the following, we describe each component in detail.

\subsubsection{EgoMoD Architecture}
\label{sec:architecture}

The architecture, visible in~\cref{fig:pipeline_diagram}, is structured into three main components:

\begin{itemize}
    \item \textbf{Video and Pose Encoding:} video clips over time $\left(t, t+n\right)$, $\mathcal{I}^{}_t,$ are processed using a frozen video foundation encoder to extract spatio-temporal features, and robot pose information $\mathbf{r}^{}_{t+n}$ is encoded using a Multi-Layer Perceptron (MLP);
    \item \textbf{Attention Module:} the two feature modalities are fused via an attention mechanism;
    \item \textbf{Map Decoder:} the resulting features are decoded by three independent heads that output predictions of $\mathcal{F}^{}_{t+T}$, $\mathcal{D}^{}_{t+T}$, and $\mathcal{E}^{}_{t+T}$.
\end{itemize}

The key idea is to combine motion cues from visual input with the global pose information over a short time $n$ to ground the observed motion and predict allocentric dynamic maps after a time $T$. The specific architectural choices are described in \cref{sec:implementation}.

\subsubsection{Computing Supervision}

Since the robot's egocentric view only covers a fraction of the scene at any given time with a limited line of sight, we leverage images from external sensors that span the whole environment to collect detections across the entire space for training. Global observations can be obtained in practice from CCTV cameras or different mobile robots covering the area of interest~\cite{drazen2013cctvdataset}.

The map is represented as a two-dimensional grid in an allocentric frame, $\mathcal{M}$, following existing MoD algorithms~\cite{molina2021robotic}. For each cell we consider $B$ motion direction bins covering $[0, 2\pi)$.
From each external camera feed, we extract detections of the dynamic entities in the scene using an object detector and depth information, obtaining a position and motion direction $(x, y, \alpha)$ in the world frame. Motion direction is estimated from centroid displacement between consecutive frames with trivial nearest-neighbor association, without the need of sophisticated data association or tracking methods.

\begin{figure}[]
    \centering
    \begin{subfigure}[]{0.11\textwidth}
         \centering
         \includegraphics[width=\textwidth]{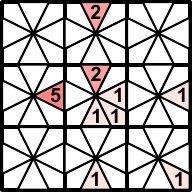}
         \caption{MoD Hist.}
         \label{fig:histogram}
     \end{subfigure}
     \begin{subfigure}[]{0.11\textwidth}
         \centering
         \includegraphics[width=\textwidth]{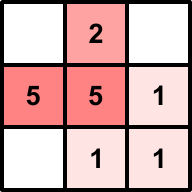}
         \caption{Flow, $\mathcal{F}$}
         \label{fig:flow}
     \end{subfigure}
     \begin{subfigure}[]{0.11\textwidth}
         \centering
         \includegraphics[width=\textwidth]{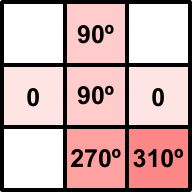}
         \caption{Direct., $\mathcal{D}$}
         \label{fig:direction}
     \end{subfigure}
    \begin{subfigure}[]{0.11\textwidth}
         \centering
         \includegraphics[width=\textwidth]{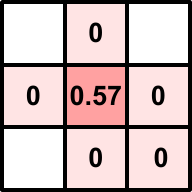}
         \caption{Entropy, $\mathcal{E}$}
         \label{fig:entropy}
     \end{subfigure}
     \caption{Example of construction of our Maps of Dynamics}
    \label{fig:maps_construction}
\end{figure}

All detections are accumulated over a time interval $T$ into the grid cells, yielding orientation histograms.
Formally, each grid cell, $c\in\mathcal{M},$ maintains a histogram of orientation counts, $h_c^{}(b), \quad b = 1, \dots, B$.
From these histograms, we derive the three descriptors:

\textbf{Flow Magnitude:} the total number of detections observed in a cell, computed as the sum of counts across all orientation bins (see~\cref{fig:flow})
\begin{equation}
f_c^{} = \sum_{b=1}^B h_c^{}(b).
\end{equation}

\textbf{Dominant Direction:} the angular bin with the maximum count, representing the most frequently observed motion direction (see~\cref{fig:direction}), 
\begin{equation}
d_c^{} = \frac{2\pi}{B} \cdot \arg\max_{b} \, h^{}_c(b).
\end{equation}
In practice, $d_c^{}$ is represented as $(\cos d_c^{}, \sin d_c^{})$ to avoid discontinuities at angular boundaries.

\textbf{Entropy}: the Shannon entropy of the orientation distribution, quantifying directional uncertainty. High entropy indicates diverse or inconsistent motion; low entropy implies coherent, directed flow (see~\cref{fig:entropy}),
\begin{equation}
e_c ^{}= - \sum_{b=1}^B \frac{h_c^{}(b)}{f^{}_c} \log \left( \frac{h_c^{}(b)}{f_c^{}} + \epsilon \right).
\end{equation}

It is important to mention that while the supervision-MoDs are aggregated over $T$ seconds of future observations, our model only uses egocentric frames for an initial short interval $n$ as input with $n \ll T$. This setup ensures that the network learns to infer future dynamics from a short initial clip.

\subsubsection{Loss Function}

We employ dedicated loss functions for each MoD descriptor. MoDs present a class imbalance between cells with observed motion and static background regions. Thus, all loss functions incorporate spatial weighting to address this imbalance. Specifically, we define a weight map $\mathcal{W}=\{w_c^{}, c\in \mathcal{M}\}$ that assigns higher weight $w^{}_{\text{valid}}$ to cells containing motion observations and lower weight $w^{}_{\text{bg}}$ to background cells.
Furthermore, pixel-wise regression loss (e.g., Huber or MSE) treats each cell independently and may lead to over-smoothed predictions that blur motion boundaries. To address this, we incorporate a gradient-based structural term that penalizes discrepancies in spatial derivatives between prediction and ground truth. 

\textbf{Flow Magnitude Loss.} Combines Huber loss with a gradient-based structural term for smooth convergence while maintaining sensitivity to spatial variations,
\begin{equation}
    \mathcal{L^{}}_{\text{flow}} = \mathcal{L}^{}_{\text{huber}}(\mathcal{F}, \hat{\mathcal{F}}, \mathcal{W}) + \lambda^{}_{\text{grad}} \cdot \mathcal{L}^{}_{\text{grad}}(\mathcal{F}, \hat{\mathcal{F}}, \mathcal{W}),
\end{equation}
where $\mathcal{F}$ and $\hat{\mathcal{F}}$ represent the ground truth and predicted flow maps respectively. The pixel-wise Huber loss is
\begin{equation}
\label{Eq:hubber}
    \mathcal{L}^{}_{\text{huber}}(\mathcal{F}, \hat{\mathcal{F}}, \mathcal{W}) = \frac{1}{|\mathcal{F}|} 
    \sum_{c\in\mathcal{F}}
    w_c^{} \cdot \mathcal{L}^{}_{\text{smooth}}(f_c^{}, \hat{f}_c^{}),
\end{equation}
with
\begin{equation}
\label{Eq:smooth}
    \mathcal{L}^{}_{\text{smooth}}(f_c^{}, \hat{f}_c^{}) = 
    \begin{cases}
    \frac{1}{2} (f_c^{} - \hat{f}_c^{})^2 & \text{if} \ |f_c^{} - \hat{f}_c^{}| \leq \beta \\
    \beta \cdot |f_c^{} - \hat{f}_c^{}| - \frac{1}{2} \beta^2 & \text{otherwise}
    \end{cases},
\end{equation}
where $\beta$ controls the transition between quadratic and linear regimes.

\textbf{Direction Loss.} For direction prediction, we predict the cosine and sine components of the angle separately and employ mean squared error on each component, combined with a gradient-based structural term,
\begin{equation}
    \mathcal{L}^{}_{\text{direction}} = \mathcal{L}^{}_{\text{angle}}(\mathcal{D}, \hat{\mathcal{D}}, \mathcal{W}) + \lambda^{}_{\text{grad}} \cdot \mathcal{L}^{}_{\text{grad}}(\mathcal{D}, \hat{\mathcal{D}}, \mathcal{W}),
\end{equation}
where $\mathcal{L}^{}_{\text{angle}}$ is the angular error loss (in terms of cosine and sine components).

\textbf{Entropy Loss.} Similar to flow magnitude, combines Huber loss with structural similarity,
\begin{equation}
    \mathcal{L}^{}_{\text{entropy}} = \mathcal{L}^{}_{\text{huber}}(\mathcal{E}, \hat{\mathcal{E}}, \mathcal{W}) + \lambda^{}_{\text{grad}} \cdot\mathcal{L}^{}_{\text{grad}}(\mathcal{E}, \hat{\mathcal{E}}, \mathcal{W}),
\end{equation}
where $\mathcal{L}^{}_{\text{huber}}$ is defined analogously to~\eqref{Eq:hubber}-\eqref{Eq:smooth} but for $\mathcal{E}$.

For all three descriptors, the gradient-based structural loss is defined as
\begin{equation}
\begin{split}
    &\mathcal{L}^{}_{\text{grad}}(\mathcal{M},\hat{\mathcal{M}},\mathcal{W}) = \\
    &\frac{1}{|\mathcal{M}|} \sum^{}_{c\in\mathcal{M}} w^{}_c \left( \| \nabla^{}_x m^{}_c - \nabla^{}_x \hat{m}^{}_c \|^2 + \| \nabla^{}_y m^{}_c - \nabla^{}_y \hat{m}^{}_c \|^2 \right)
    \end{split}
\end{equation}
where $\nabla^{}_x$ and $\nabla^{}_y$ denote the horizontal and vertical gradients computed via finite differences. For direction, $\mathcal{L}^{}_{\text{grad}}$ is also computed on the cosine and sine components separately.

\section{Implementation Details}
\label{sec:implementation}

\textbf{Video and Pose Encoding.} We use V-JEPA2 (ViT-G variant with 384$\times$384 input resolution) as the visual backbone to extract spatio-temporal features from egocentric video clips~\cite{assran2025v}. Each input sequence consists of $f = 8$ RGB frames during time $n$. The V-JEPA encoder processes these frames and produces $N=2304$ patch-level feature vectors $\mathbf{V}$, each with dimension $D=1408$, arranged in a $48 \times 48$ grid. The V-JEPA encoder weights remain frozen during training; only the attention modulen and decoders are trained.

The robot pose is represented as a 7-dimensional vector $\textbf{r} = (x, y, z, q_x, q_y, q_z, q_w)$, where $(x, y, z)$ is the 3D position and $(q_x, q_y, q_z, q_w)$ is the orientation quaternion. The pose encoder network consists of a two-layer MLP ($7{\to}128{\to}64$) with ReLU activation, and layer normalization, producing a 64-dimensional pose embedding. This embedding is then linearly projected to match the visual feature dimension $D$ for integration with the Transformer, yielding $\mathbf{z}^{}_p \in \mathbb{R}^{D}_{}$.

\textbf{Attention Module.} The pose embedding is treated as a pose token and prepended to the sequence of patch features. The resulting sequence $[\mathbf{z}^{}_p; \mathbf{V}] \in \mathbb{R}^{(1+N) \times D}$ is processed by a Transformer encoder with 2 layers and 8 attention heads (feed-forward dimension of 512, dropout of~0.2). Self-attention allows all patches to attend to the pose token and to each other, enabling the network to learn pose-conditioned spatial relationships. After processing, the pose token is discarded, and only the patch outputs are retained.

\textbf{Map Decoder.} The $N$ patch outputs are linearly projected from $D$ to 256 dimensions and reshaped to a $48 \times 48 \times 256$ spatial tensor. A convolutional decoder with progressive bilinear upsampling then maps this representation to the target map resolution. The decoder produces the final MoD prediction $\hat{\mathcal{M}}^{}_{t+T} \in \mathbb{R}^{H \times W}$, where the number of output channels and final activation depend on the target map type:

\begin{itemize}
    \item \textbf{Flow and Entropy:} 1 output channel with sigmoid activation (values in $[0, 1]$)
    \item \textbf{Direction:} 2 output channels with tanh activation (cosine and sine components)
\end{itemize}

\textbf{Training Configuration.} The model is trained using the AdamW optimizer with a polynomial learning rate schedule. \cref{tab:parameters} summarizes the key hyperparameters.
\begin{table}[t]
    \centering
    \caption{Model, training, and loss hyperparameters.}
    \resizebox{\linewidth}{!}{
    \begin{tabular}{lclc}
        \toprule
        \textbf{Parameter} & \textbf{Value} & \textbf{Parameter} & \textbf{Value}\\
        \midrule
        \multicolumn{2}{l}{\textit{Map of Dynamics}} &  \multicolumn{2}{l}{\textit{Regularization}} \\
        Direction bins $B$ & 8 &  Input (patch) dropout & 0.2 \\
        Temporal horizon $T$ & 10 / 20\,s &  &  \\
        Grid cell size & 0.30\,m & & \\
        \midrule
        \multicolumn{2}{l}{\textit{Loss Weights}} & \multicolumn{2}{l}{\textit{Optimization}} \\
        Valid region weight $w_{\text{valid}}$ & 5.0 & Learning rate & $1 \times 10^{-4}$ \\
        Background weight $w_{\text{bg}}$ & 0.95 & Weight decay & $1 \times 10^{-4}$ \\
        Gradient loss weight $\lambda_{\text{grad}}$ & 1.0 &  Batch size & 8 \\
        Huber threshold $\beta$ & 0.1 & Number of epochs & 50 \\
        \bottomrule
    \end{tabular}   
}

    \label{tab:parameters}
\end{table}

\subsection{Data Augmentation}

To improve generalization, we apply the following augmentations during training:
\begin{itemize}
    \item \textbf{Feature noise:} Gaussian noise with $\sigma=0.1$ multiplied element-wise to the V-JEPA features
    \item \textbf{Feature dropout:} Random dropout of individual feature dimensions with probability 0.01
    \item \textbf{Pose translation:} Random translation in $X$ and $Y$ up to $\pm 0.2$\,m
    \item \textbf{Pose rotation:} Random rotation around the vertical axis up to $\pm 5^\circ$
\end{itemize}

\section{Experimental Evaluation}

\begin{figure}
    \centering
    \begin{subfigure}[]{0.25\textwidth}
        \centering
        \includegraphics[width=\textwidth]{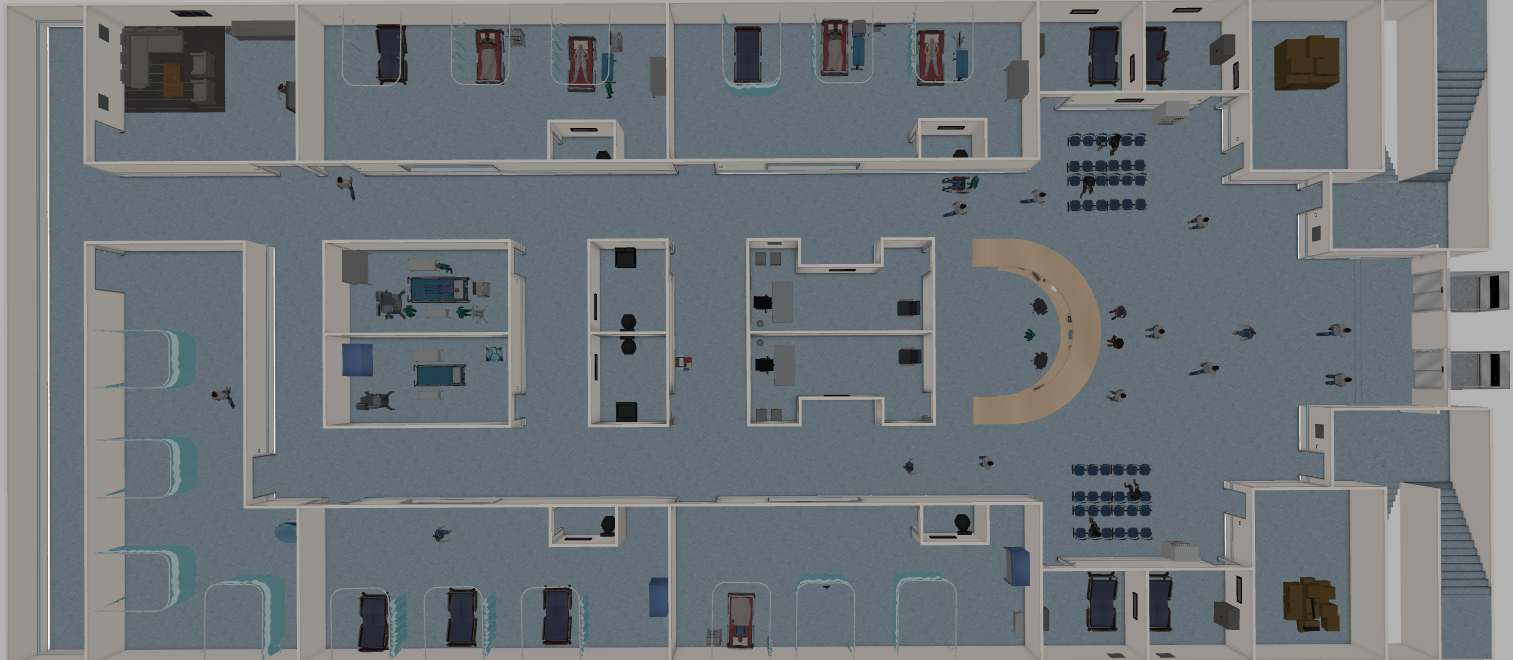}
        \caption{Top-down view}
        \label{fig:hospital_simulation_top}
    \end{subfigure}
    \begin{subfigure}[]{0.23\textwidth}
         \centering
         \includegraphics[width=\textwidth, trim={5cm 4cm 2cm 4cm},clip]{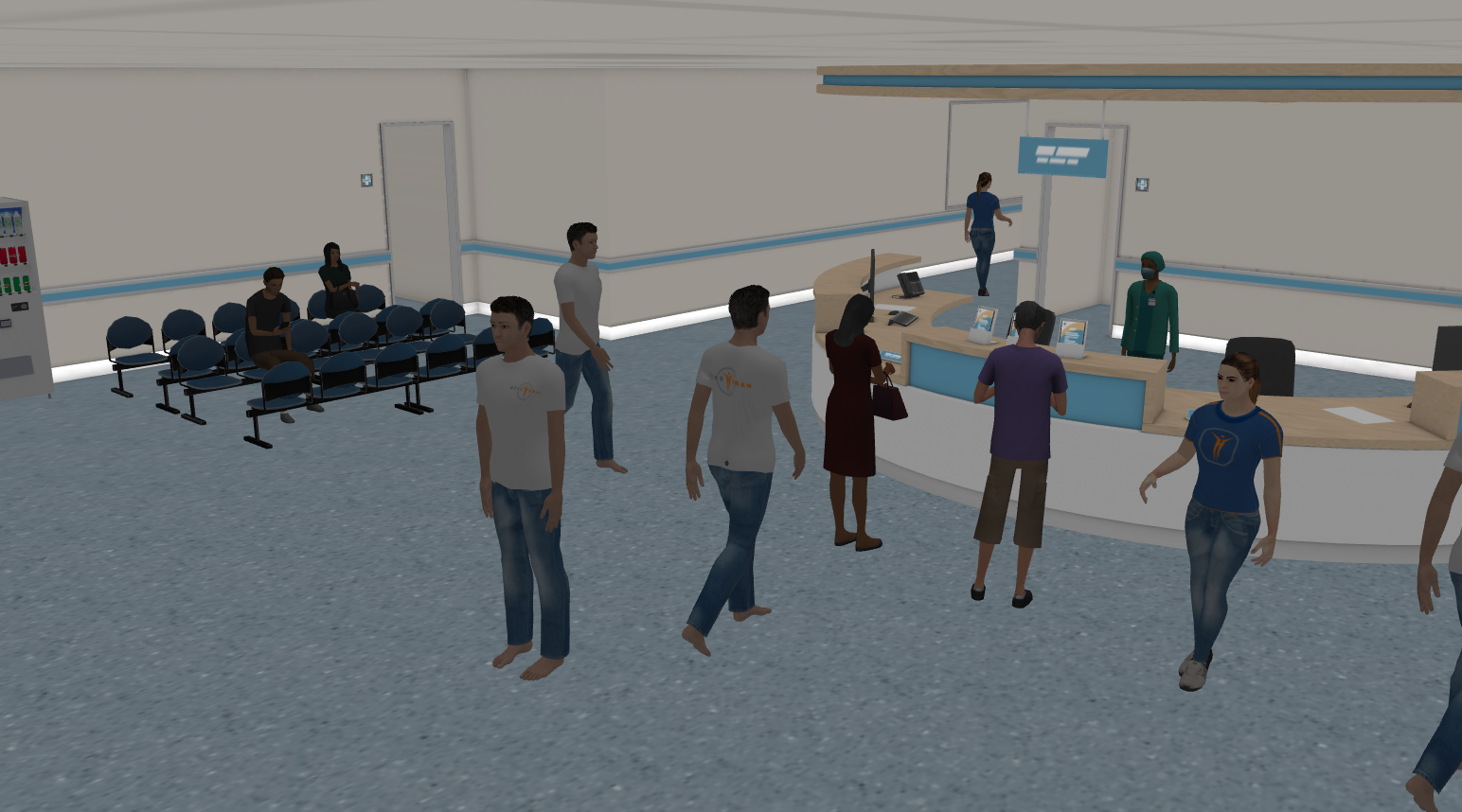}
         \caption{Perspective view}
         \label{fig:hospital_simulation_bottom}
     \end{subfigure}
    \caption{Overview of the hospital simulation environment shown from a top-down view (left) and a perspective view (right).}
    \label{fig:hospital_simulation}
\end{figure}

\subsection{Simulation Environments}
\label{sec:simulation}

The training environment used for quantitative evaluation is a hospital scenario built in Gazebo using AWS RoboMaker Hospital World, combined with a customized version of PedSim simulator~\cite{xie2021towards} extended to include this hospital setting (see \cref{fig:hospital_simulation}). Additionally, the agent behaviors were modified to define movement patterns and queuing behaviors at key locations. The hospital measures approximately 26\,m by 56\,m and contains 14 scenes designed for this study. Each scene displays diverse motions with a varying numbers of agents (10--15). The agents move following realistic low-level human dynamics and represent doctors and patients performing activities such as visiting beds and waiting at reception areas. We release the code to run our simulator and scene descriptions.

\subsection{Real MoD computation}
Our first experiment assesses the capacity to construct a MoD by fusing detections from multiple video sources in the simulation environment.
Observations are collected by three static cameras. Real detections of the people in the scene are obtained using YOLOv11~\cite{yolo11_ultralytics}.
Each detection is lifted into 3D space and associated with a motion direction estimate based on short-term temporal tracking.

To assess the feasibility of using real-world detections for map generation, we compare the MoD generated with the real perception pipeline against that generated from simulator-provided detections. 
We adopt a distributional comparison strategy across map types (flow, direction, entropy). We quantify differences between maps derived from simulator-provided and YOLO-based detections using: (i) Jensen–Shannon (JS) divergence as a bounded measure of distributional dissimilarity, (ii) Bhattacharyya distance to capture overlap between distributions, and (iii) for direction maps, angular similarity to assess alignment in predominant motion directions.

The results indicate that flow maps exhibit the lowest distributional discrepancy (JS\,=\,0.42, Bhattacharyya\,=\,0.23) among the evaluated map types, suggesting that YOLO detections preserve the spatial distribution of motion intensity despite reduced coverage. For direction maps, JS divergence (0.40) and Bhattacharyya distance (0.24) indicate moderate divergence, with an angular similarity of 0.47 reflecting partial alignment in dominant motion patterns. Entropy maps display the highest discrepancy (JS\,=\,0.54, Bhattacharyya\,=\,0.45), consistent with their sensitivity to fragmented or missing detections.

\subsection{Dataset}

For quantitative evaluation (\cref{tab:baseline_comparison}), we use the simulator described in \cref{sec:simulation}. A mobile robot explores the environment along predefined trajectories while capturing egocentric RGB videos and 6-DoF pose information. From this data, we generate ground-truth motion maps by aggregating simulator-provided person detections as detailed in \cref{sec:histogram_mod}. In total, the dataset contains 2500 usable training sequences, each consisting of 8 RGB frames and corresponding robot poses. The dataset is partitioned into 80\% for training and 20\% for validation. For testing, we leave out an entire scene from training to evaluate generalization.
We verify that it constitutes an out-of-distribution test by computing the per-cell standard deviation of the GT flow map across frames, and the mean correlation between individual frames and the scene's own mean map. The test scene has the highest within-scene flow variation (cell-wise std: 0.197) and the lowest frame-to-mean correlation (0.536) across all scenes, confirming that its per-frame dynamics cannot be resolved by memorizing a single representative map.
We evaluate at two prediction horizons, 10\,s and 20\,s, to demonstrate the capacity of our model to operate at different temporal MoD scales.

\begin{table*}[]
    \centering
    \caption{Quantitative comparison on the hospital simulation environment. EgoMoD outperforms both baselines on flow metrics and achieves consistent improvements across evaluation scopes and horizons. 
    Local-StefMap stores only within-FOV predictions, making global ECR unavailable~($-$).}
    \label{tab:baseline_comparison}
    \begin{tabular}{@{}cclccccccccc@{}}
        \toprule
        \multicolumn{1}{c}{\multirow{2}{*}{Horizon}}
        & \multicolumn{1}{c}{\multirow{2}{*}{Observability}}
        & \multicolumn{1}{c}{\multirow{2}{*}{Method}}
        & \multicolumn{3}{c}{Flow}
        & \multicolumn{3}{c}{Entropy}
        & \multicolumn{2}{c}{Direction}
        & \multicolumn{1}{c}{ECR} \\

        & &
        & $\downarrow$ MSE & $\downarrow$ MAE & $\uparrow$ SSIM
        & $\downarrow$ MSE & $\downarrow$ MAE & $\uparrow$ SSIM
        & $\uparrow$ Acc. & $\uparrow$ IoU
        & $\uparrow$ ECR \\
        \midrule

        \multirow{6}{*}{10 s}
        & \multirow{3}{*}{Local}
        & Mean-MoD
        & 59.26 & 5.38 & 0.52
        & 19.26 & 4.27 & 0.23
        & \textbf{0.70} & \textbf{0.23}
        & 0.00 \\

        &
        & Local-StefMap~\cite{molina2021robotic}
        & 204.86 & 8.83 & 0.31
        & 30.66 & 5.33 & 0.19
        & 0.29 & 0.04
        & $-$ \\

        &
        & EgoMoD (ours)
        & \textbf{48.43} & \textbf{4.55} & \textbf{0.66}
        & \textbf{13.00} & \textbf{2.86} & \textbf{0.30}
        & 0.44 & 0.11
        & \textbf{0.23} \\

        \cmidrule(lr){2-12}

        & \multirow{3}{*}{Global}
        & Mean-MoD
        & 63.94 & 5.65 & 0.84
        & 19.26 & 4.27 & 0.73
        & \textbf{0.69} & \textbf{0.26}
        & 0.00 \\

        &
        & Local-StefMap~\cite{molina2021robotic}
        & 185.24 & 8.35 & 0.81
        & 33.3 & 5.74 & 0.73
        & 0.24 & 0.02
        & $-$ \\

        &
        & EgoMoD (ours)
        & \textbf{53.79} & \textbf{4.78} & \textbf{0.89}
        & \textbf{13.22} & \textbf{2.92} & \textbf{0.76}
        & 0.42 & 0.10
        & \textbf{0.21} \\

        \midrule

        \multirow{6}{*}{20 s}
        & \multirow{3}{*}{Local}
        & Mean-MoD
        & 48.47 & 4.76 & 0.64
        & 12.55 & 3.27 & 0.26
        & \textbf{0.72} & \textbf{0.26}
        & 0.07 \\

        &
        & Local-StefMap~\cite{molina2021robotic}
        & 220.03 & 9.59 & 0.26
        & 30.02 & 5.29 & 0.17
        & 0.25 & 0.03
        & $-$ \\

        &
        & EgoMoD (ours)
        & \textbf{35.72} & \textbf{3.82} & \textbf{0.79}
        & \textbf{6.33} & \textbf{1.67} & \textbf{0.38}
        & 0.58 & 0.15
        & \textbf{0.44} \\

        \cmidrule(lr){2-12}

        & \multirow{3}{*}{Global}
        & Mean-MoD
        & 59.16 & 5.20 & 0.84
        & 12.56 & 3.29 & 0.65
        & \textbf{0.73} & \textbf{0.29}
        & 0.05 \\

        &
        & Local-StefMap~\cite{molina2021robotic}
        & 213.91 & 9.19 & 0.71
        & 32.16 & 5.62 & 0.63
        & 0.24 & 0.02
        & $-$ \\

        &
        & EgoMoD (ours)
        & \textbf{46.92} & \textbf{4.22} & \textbf{0.9}
        & \textbf{5.38} & \textbf{1.55} & \textbf{0.71}
        & 0.58 & 0.16
        & \textbf{0.44} \\

        \bottomrule
    \end{tabular}
\end{table*}

\subsection{Evaluation Metrics}

We evaluate our method using metrics appropriate for each map type:

\textbf{Flow and Entropy Maps:}
\begin{itemize}
    \item \textbf{MSE}: Mean Squared Error between predicted and ground truth maps
    \item \textbf{MAE}: Mean Absolute Error for pixel-wise comparison
    \item \textbf{SSIM}: Structural Similarity Index to assess spatial coherence
\end{itemize}

\textbf{Direction Maps:}
\begin{itemize}
    \item \textbf{Accuracy}: Classification accuracy when direction is discretized into bins
    \item \textbf{IoU}: Intersection over Union for discretized direction classes
\end{itemize}

\textbf{ECR (Entropy-Calibrated direction hit Rate):}
To jointly evaluate whether direction predictions are practically actionable, we introduced a new metric reporting the fraction of valid cells where predicted entropy is below one third of its maximum value \emph{and} the dominant direction is within $45^\circ$ of ground truth. A high ECR indicates that the model provides confident directional guidance only when that guidance is likely correct.

\subsection{Quantitative Evaluation}

Owing to the lack of directly comparable approaches in the literature, we define two baselines.
\textit{Mean-MoD} is a parameter-free reference that predicts a single constant map per head, computed as the per-cell training average with no input (arithmetic mean for flow and entropy; per-cell circular mean for direction). \textit{Local-StefMap}, based on the work in ~\cite{molina2021robotic}, constructs motion maps using people detections from a YOLOv11 model applied to egocentric RGB-D videos. Following the model outlined in~\cref{sec:histogram_mod}, \textit{Local-StefMap} accumulates orientation histograms in a spatial grid over time, but only within the robot’s current field of view. Unlike our proposed method, which predicts motion patterns across the entire environment, \textit{Local-StefMap} computes dynamics strictly in regions currently visible to the robot and does not attempt to infer unobserved areas.

To ensure fair evaluation, we compare all methods under two conditions: (i) \textit{local observability}, where predictions are cropped to the robot’s field of view, and (ii) \textit{global observability}, where predictions cover the full spatial extent of the environment. In addition, we evaluate performance across two prediction horizons, 10s and 20s. We also report ground truth motion maps in both local and global forms to contextualize each method’s performance.

The results in \cref{tab:baseline_comparison} demonstrate that EgoMoD outperforms both baselines on flow and entropy metrics across all evaluation scopes and time horizons. EgoMoD achieves lower MSE and higher SSIM, indicating both improved accuracy and better preservation of spatial structure. For direction, EgoMoD surpasses Local-StefMap in classification accuracy (0.42 vs.\ 0.24 at 10\,s global), confirming that the model successfully captures dominant motion directions in the scene.
EgoMoD achieves ECR~$= 0.21$~(10\,s, global) and $0.44$~(20\,s, global). Local-StefMap achieves slightly higher ECR at 10\,s ($0.23$ global) because it restricts predictions to directly-observed cells, which are well-calibrated by construction; EgoMoD’s ECR advantage doubles at 20\,s ($0.44$ vs $0.23$), where global prediction capability enables actionable guidance across the full scene. Mean-MoD’s near-zero ECR ($0.00$ and $0.05$) confirms that without input conditioning no cell clears the joint confidence-and-accuracy bar.
We measure that the network inference time is below 200 ms in our experiments.

\subsection{Input Ablation Study}

\begin{table}[]
\centering
\caption{Ablation study on the model input.}
\label{tab:flow-perturbation}
\small
\setlength{\tabcolsep}{4pt}
\begin{tabular}{@{}lcccc@{}}
    \toprule
    \multirow{2}{*}{\textbf{Ablation}}
    & \multicolumn{1}{c}{Flow}
    & \multicolumn{1}{c}{Entropy}
    & \multicolumn{1}{c}{Direction}
    & \multicolumn{1}{c}{ECR} \\
    & $\downarrow$ MAE 
    & $\downarrow$ MAE 
    & $\uparrow$ Acc.
    & $\uparrow$ ECR \\
    \midrule
    baseline
        & \textbf{4.78} 
        & 2.92
        & \textbf{0.42}
        & 0.21 \\
    vision: no\_motion
        & 4.87 
        & 3.15 
        & 0.41
        & 0.19 \\
    vision: motion
        & 4.94 
        & 2.67 
        & 0.36
        & 0.21 \\
    vision: static
        & 5.32 
        & \textbf{2.28} 
        & 0.36
        & \textbf{0.22} \\
    vision: OOD
        & 5.75
        & 3.91 
        & 0.24
        & 0.06 \\
    pose: random
        & 4.79 
        & 2.92 
        & \textbf{0.42}
        & 0.21 \\
    pose: initial
        & 4.79 
        & 2.92 
        & \textbf{0.42}
        & 0.21 \\
    \bottomrule
\end{tabular}
\end{table}

\begin{figure*}[]
    \centering
    \includegraphics[width=0.8\textwidth]{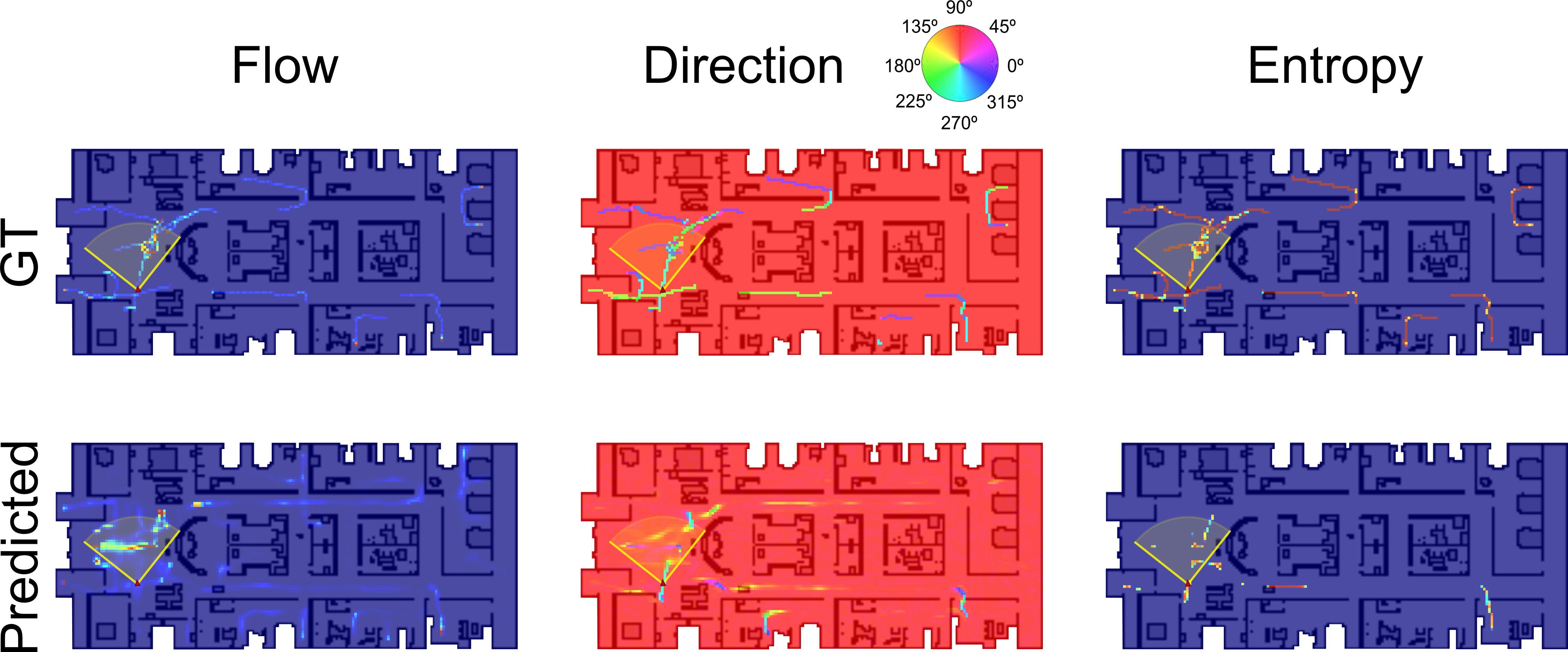}
    \caption{Qualitative results showing predicted flow, entropy, and direction maps from our model on test scenes, compared to ground truth.}
    \label{fig:qualitative_results}
    \vspace{-5pt}
\end{figure*}

\begin{figure}
    \centering
    \includegraphics[width=0.8\linewidth]{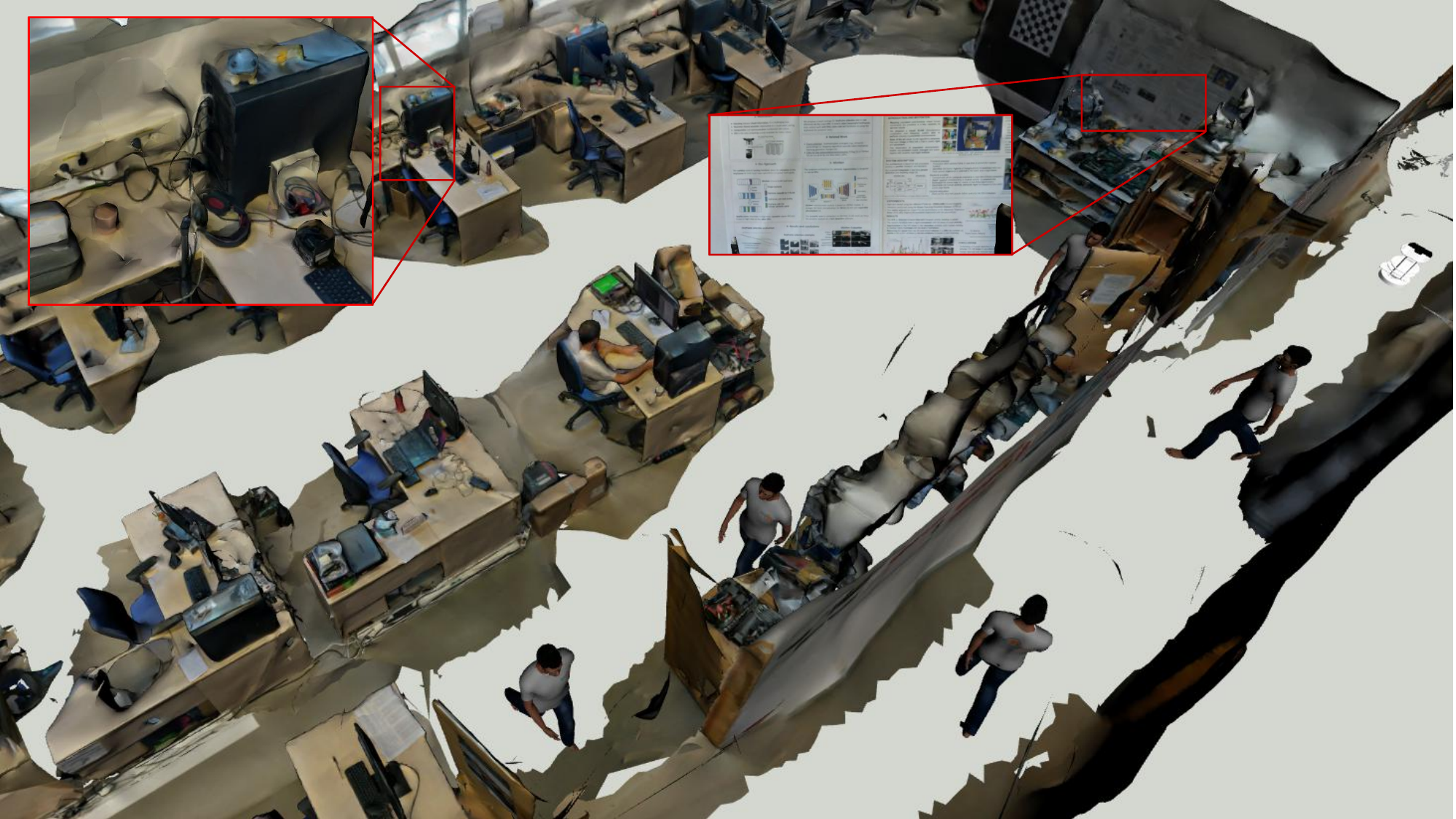}
    \caption{Photorealistic office simulated environment}
    \label{fig:office_simulation}
\end{figure}

To assess what the trained model actually reads from its inputs, we run the full EgoMoD model on the held-out scene with controlled input substitutions (\cref{tab:flow-perturbation}). Vision substitutions replace the V-JEPA patch tensor with a frame without pedestrians (\emph{no\_motion}), a frame with multiple visible pedestrians (\emph{motion}), a degenerate static clip (\emph{static}), or an out-of-distribution real-world clip (\emph{OOD}). Pose substitutions replace the 7-vector with a random or fixed-initial pose.

Two findings are consistent across all substitution types. First, replacing the visual input with an OOD clip degrades global flow MAE by $+0.022$, substantially larger than any within-distribution substitution, confirming that the visual input is actively read by the trained model and is sensitive to distributional shift. Second, pose substitutions cause changes of less than $0.001$ in all flow metrics, confirming that the model does not condition its predictions on the exact pose at inference time, reducing the risk of memorizing pose-to-map correspondences. Together, these results show that the visual stream is engaged and responds to input content, while pose provides a calibration signal that decouples from per-sample memorization.

\subsection{Real-World Experiments}

To evaluate real-world applicability, we create a second environment as a photorealistic virtual replica of our university office, comprising a main room and an adjacent corridor (see \cref{fig:office_simulation}). This environment features 6 agents following three structured motion patterns: an L-shaped trajectory through a doorway connecting the room and corridor, and two rectangular loops, one confined to the corridor and another traversing between the corridor and main room through two doors. This controlled setup was specifically designed to minimize the sim-to-real gap: after training in simulation, we deploy the model on physical robots in the same real-world office with identical motion patterns.

We conduct a small-scale deployment of the trained model in the physical university office environment that served as the basis for the office simulation. A mobile robot equipped with an RGB camera and localization system records egocentric video while 6 participants move through the space following similar motion patterns to those used in simulation.

\begin{figure}
    \centering
    \begin{subfigure}[]{0.2\textwidth}
        \centering
        \includegraphics[width=\textwidth]{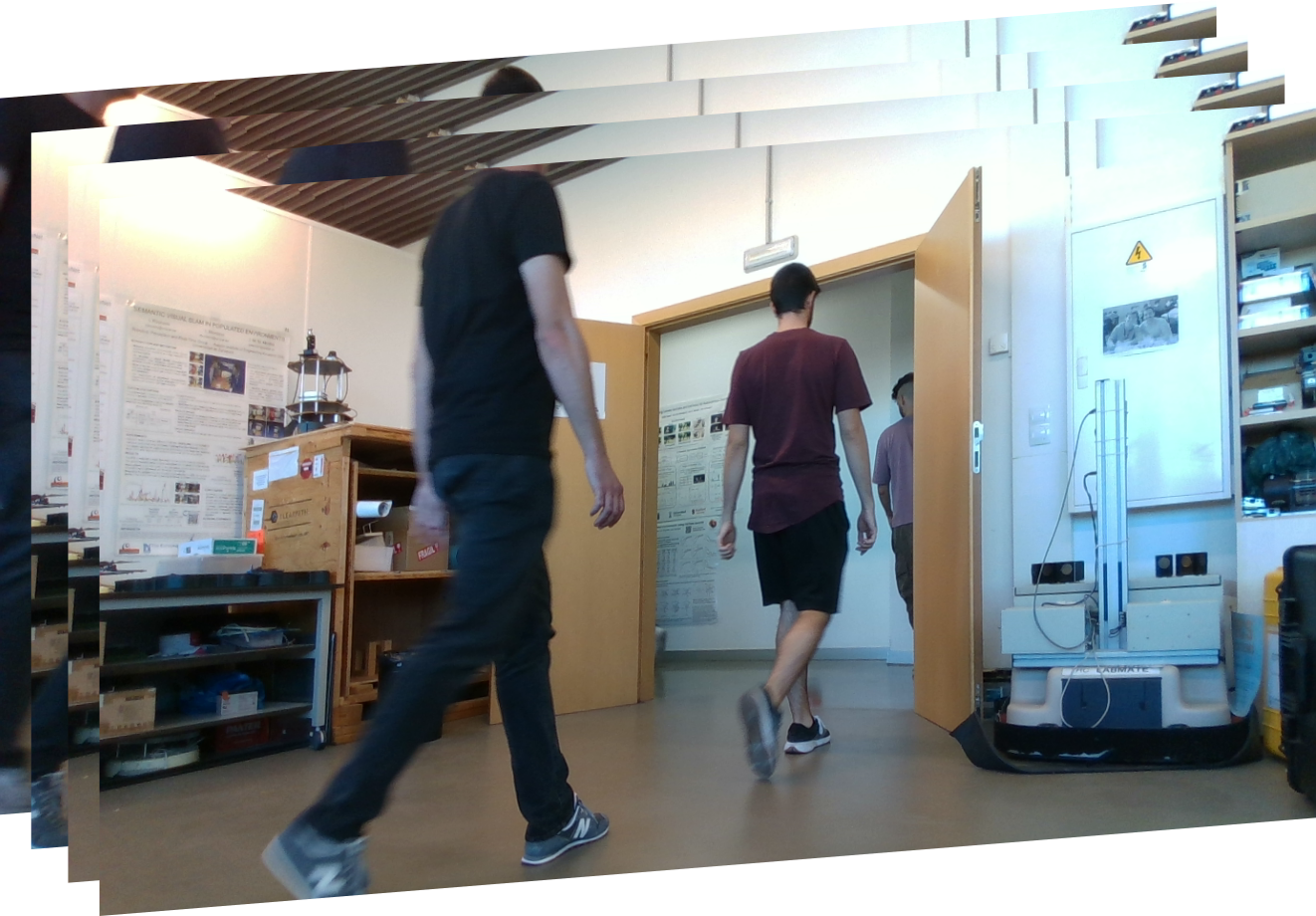}
        \caption{Perspective view}
        \label{fig:perspective_view}
    \end{subfigure}
    \begin{subfigure}[]{0.24\textwidth}
        \centering
        \includegraphics[width=\textwidth]{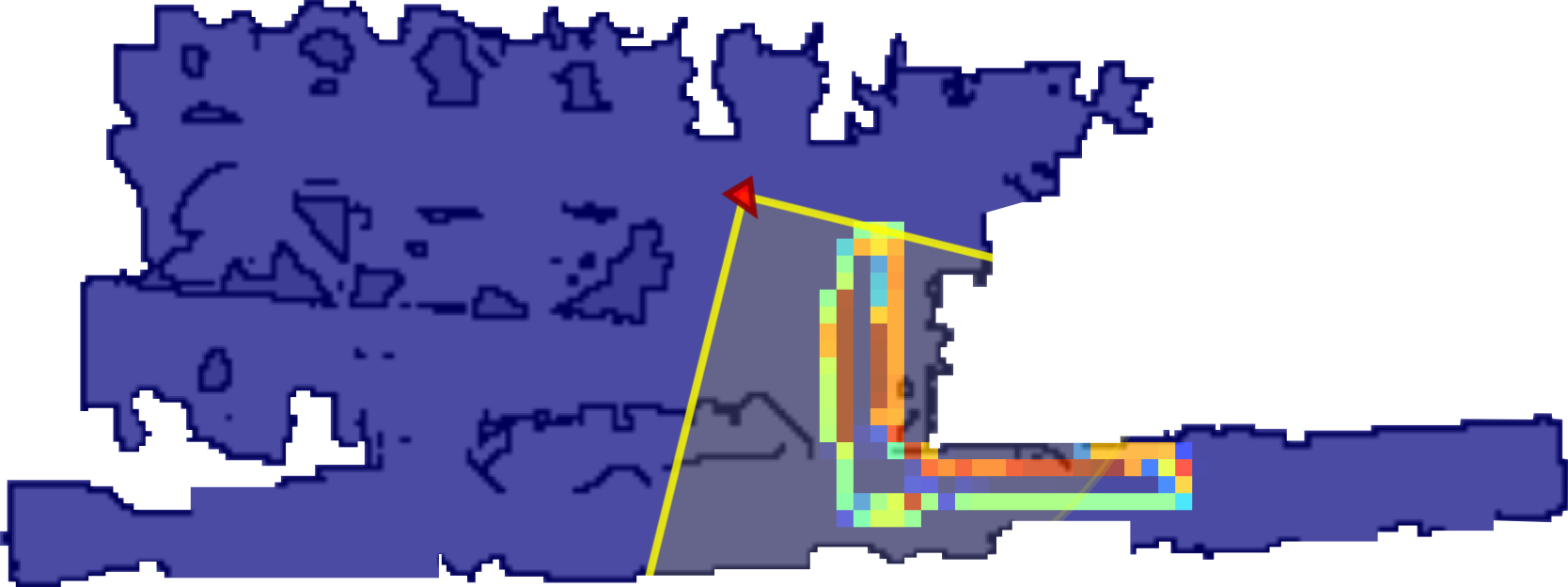}
        \caption{Entropy Prediction}
        \label{fig:real_pred_entropy}
    \end{subfigure}
    \begin{subfigure}[]{0.24\textwidth}
        \centering
        \includegraphics[width=\textwidth]{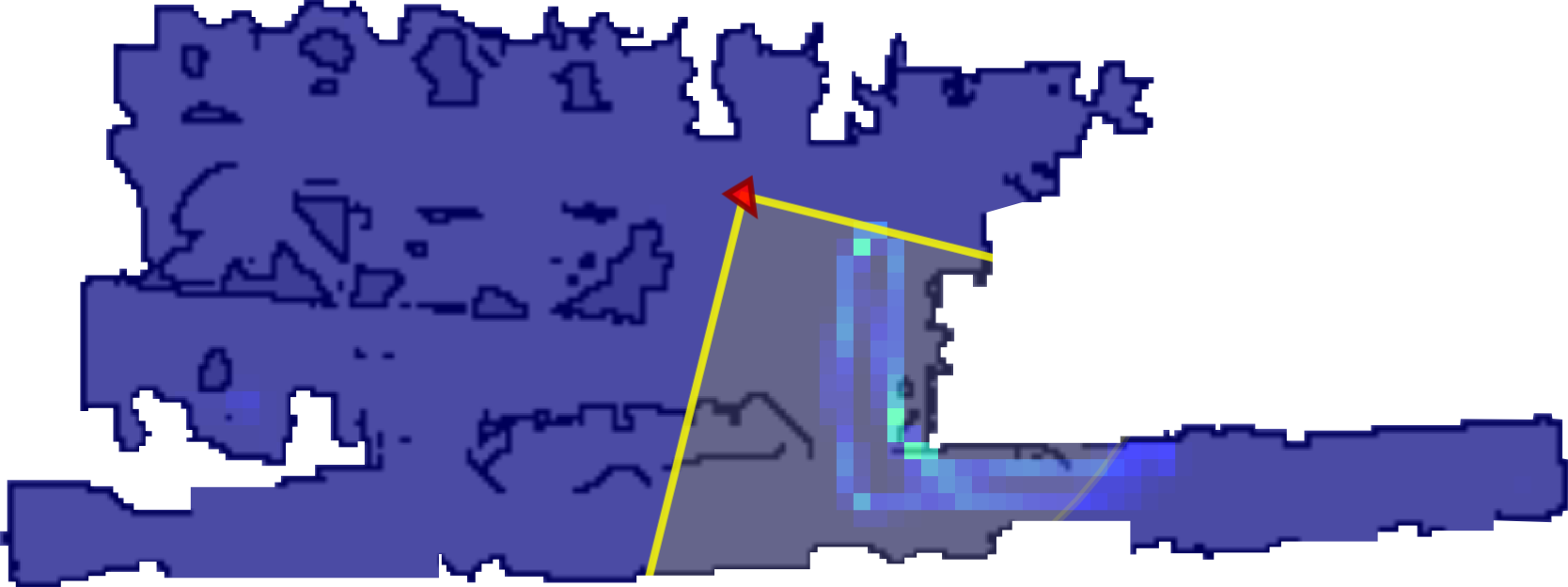}
        \caption{Flow Prediction}
        \label{fig:real_pred_flow}
    \end{subfigure}
    \begin{subfigure}[]{0.24\textwidth}
        \centering
        \includegraphics[width=\textwidth]{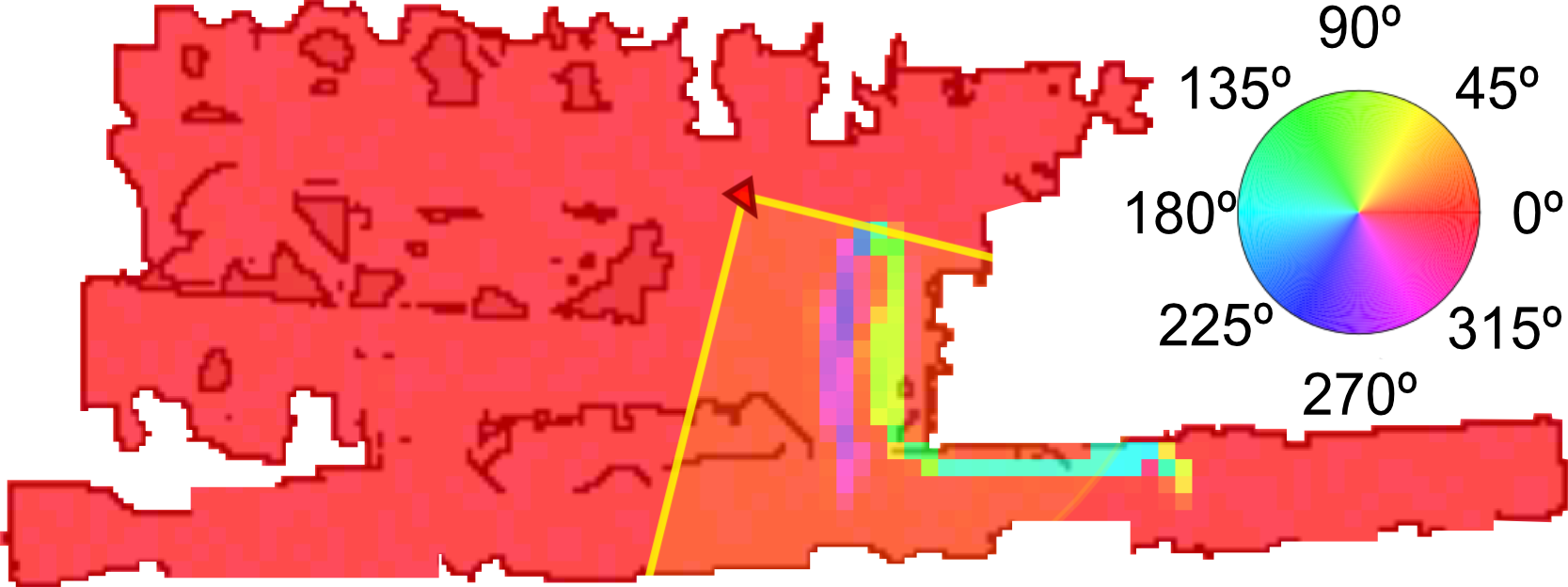}
        \caption{Direction Prediction}
        \label{fig:real_pred_direction}
    \end{subfigure}
    \caption{Qualitative results for the MoD predicted by our network on a real video clip.}
    \label{fig:real_world}
\end{figure}

\begin{figure}
    \centering
    \begin{subfigure}[]{0.24\textwidth}
        \centering
        \includegraphics[width=\textwidth]{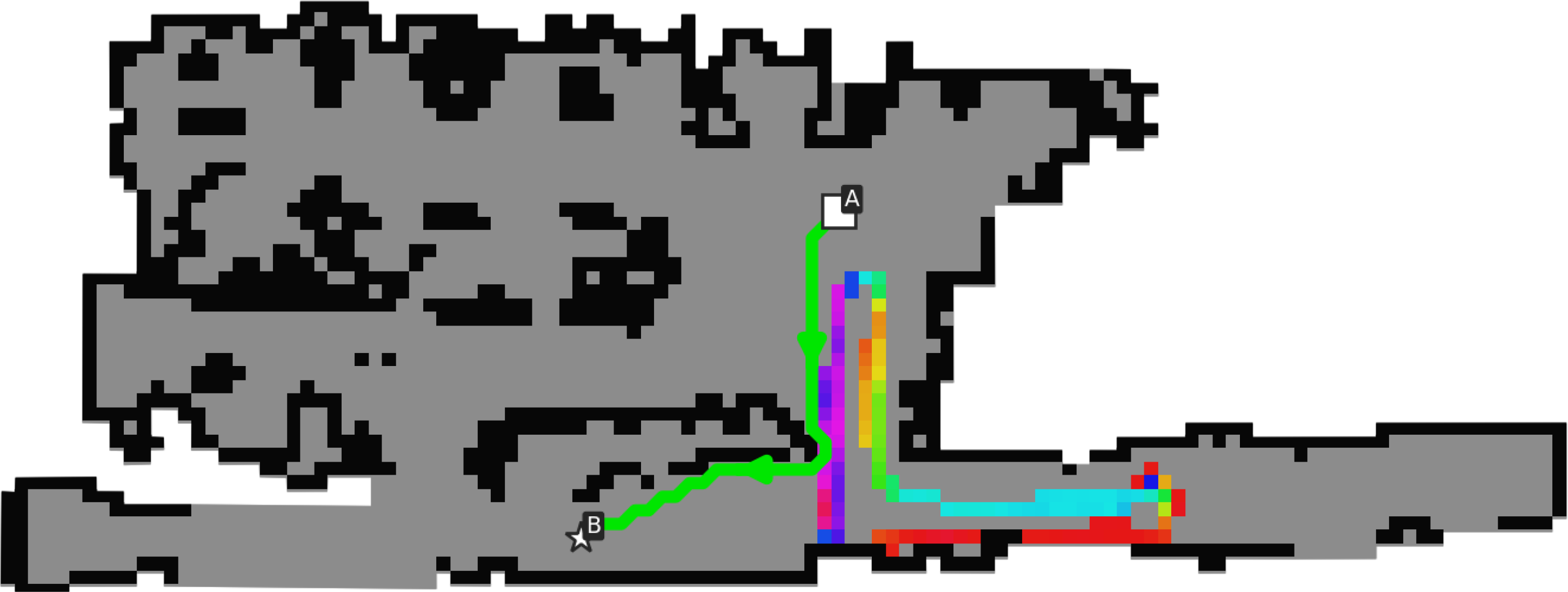}
        \caption{A$\rightarrow$B with flow}
        \label{fig:real_planner_short}
    \end{subfigure}
    \begin{subfigure}[]{0.24\textwidth}
        \centering
        \includegraphics[width=\textwidth]{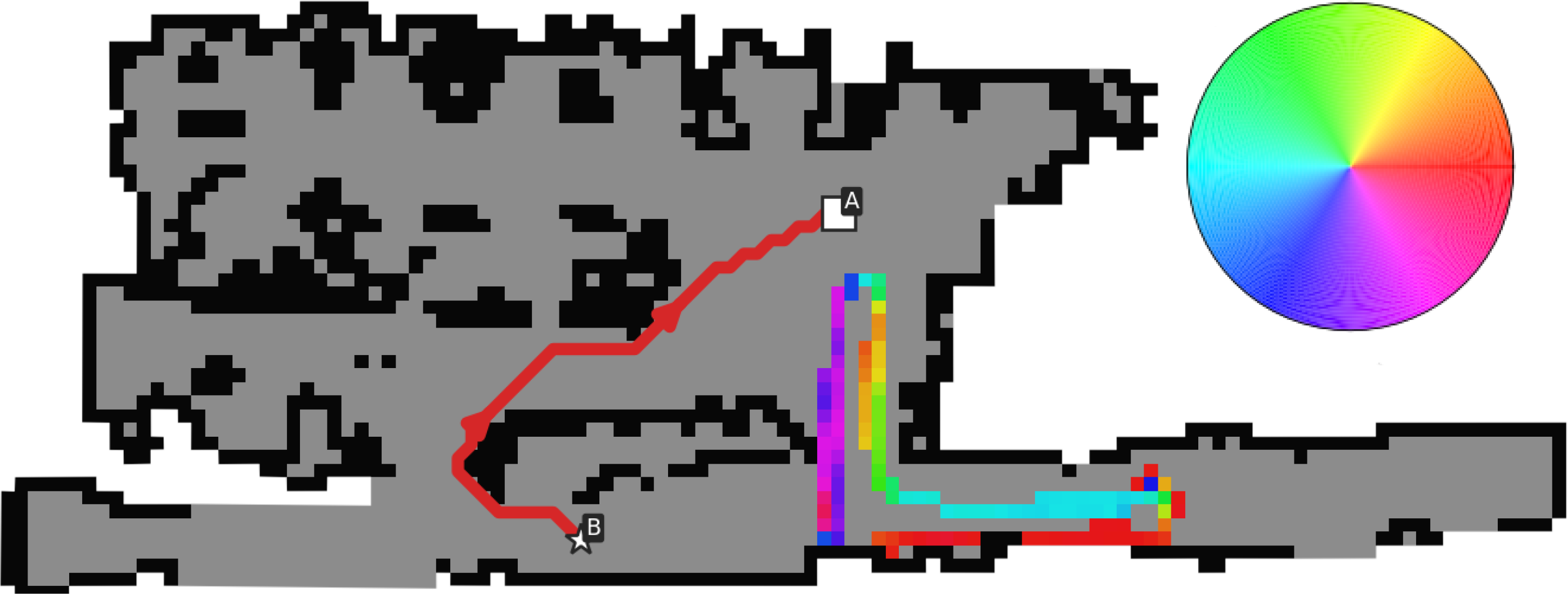}
        \caption{B$\rightarrow$A against flow}
        \label{fig:real_planner_long}
    \end{subfigure}
    \caption{ECR-based A* planning on EgoMoD's predicted direction map. Direction- aware costs route A$\rightarrow$B with the dominant flow (left) and redirect B$\rightarrow$A through an alternative path (right) to avoid moving against it.}
    \label{fig:real_planner}
\end{figure}

Since ground-truth motion maps are unavailable in real settings, we qualitatively visualize EgoMoD's forecasting ability (\cref{fig:real_world}). The results suggest that our model generalizes to real sensor inputs and produces coherent spatial dynamics predictions, even in the presence of perceptual noise and partial observations. To illustrate downstream utility, \cref{fig:real_planner} shows an ECR-based $A^*_{}$ planner operating on the predicted direction map. The planner exhibits asymmetric behavior depending on the travel direction: in A$\rightarrow$B it follows the dominant pedestrian flow and selects a short route, whereas in B$\rightarrow$A the higher cost of moving against the flow leads it to take a longer alternative path through less occupied regions. This demonstrates that EgoMoD predictions provide actionable directional guidance for flow-aware navigation. Additional qualitative results are provided in the supplementary video.

\section{Discussion \& Limitations}

Despite the encouraging experimental results, our approach has some limitations.
First, EgoMoD aimed at modeling individual environments with diverse motion patterns. Generalization across scenes and different temporal horizons requires scaling the training data.
Second, the scarcity of such training data remains a fundamental bottleneck and the main reason for leveraging simulation data in this work.
No existing egocentric robotics dataset provides the combination of RGB video, accurate localization, and dense pedestrian annotations required to construct supervision. The availability of more diverse scenes, integrating real human dynamics across different horizons and times of the day, has the potential to improve EgoMoD towards improved generalization.

\section{Conclusions}

We presented EgoMoD, a novel learning-based framework capable of predicting global Maps of Dynamics (MoDs) from short egocentric video sequences. Unlike traditional methods, EgoMoD leverages the semantic reasoning capabilities of a video encoder fused with robot localization to infer allocentric crowd dynamics.
To enable learning of global information from local observations, we have used external cameras during training to obtain ground truth MoDs and use them as supervision to our model.
Experiments in simulated environments and ablation studies show that EgoMoD outperforms the baselines across all evaluation scopes and horizons, infers motion in unobserved regions and uses input data meaningfully. We validated the approach in real-world scenarios, demonstrating that the system generalizes to real sensor data and produces coherent predictions without complex sim-to-real adaptation. 
Future work will focus on extending the representation to generalize to different scenes and time horizons by extending the input to our model and scaling the training data.

\bibliographystyle{ieeetr}
\bibliography{bibliography}

\end{document}